# Pose Matters: Pose Guided Graph Attention Network for Person Re-identification

Zhijun He[a,b], Hongbo Zhao*[a,b], Wenquan Feng[a]

[a] School of Electronics and Information Engineering, Beihang University, Beijing, China 100191;
[b] Hefei Innovation Research Institute of Beihang University, Hefei, China 230012
*bhzhb@buaa.edu.cn

## ABSTRACT

Person re-identification (reID) aims at retrieving a person from images captured by different cameras. For deep-learning-based reID methods, it has been proved that using local features together with global feature could help to give robust representation for person retrieval. Human pose information could provide the locations of human skeleton to effectively guide the network to pay more attention on these key areas and could also help to reduce the noise distractions from background or occlusion. However, methods proposed by previous pose-based works might not be able to fully exploit the benefits of pose information and few of them take into consideration the different contributions of separate local features. In this paper, we propose a pose guided graph attention network, a multi-branch architecture consisting of one branch for global feature, one branch for mid-granular body features and one branch for fine-granular key point features. We use a pre-trained pose estimator to generate the key-point heatmaps for local feature learning and carefully design a graph attention convolution layer to re-assign the contribution weights of extracted local features by modeling the similarities relations. Experiment results demonstrate the effectiveness of our approach on discriminative feature learning and we show that our model achieves state-of-the-art performances on several mainstream evaluation datasets. We also conduct a plenty of ablation studies and design different kinds of comparison experiments for our network to prove its effectiveness and robustness, including occluded experiments and cross-domain tests.

# 1. INTRODUCTION

Person re-identification (reID), aiming to retrieve a specific person across different times and cameras, has been getting more and more attention from both industry and academia. With the great advances in deep convolutional neural network over the last decade, person representation could be extracted in a much more discriminative and robust way, which pushed the performance of reID to a new level. Many existing researches [1,2,3,4,5,6] have shown that CNN-based methods could achieve better identification rates and mean average precision compared with traditional hand-crafted methods. In recent years, more and more studies tend to descript person identities based on local feature mining, which has been confirmed to be effective in many previous works [7,8,9,10,11]. Learning part-level features could help to filtering the useless information and make the networks concentrate more on identity-relevant areas. In this paper, we aim to improve the performance and robustness of person representation by using pose information and Graph Convolutional Network (GCN).

The motivation of this work is threefold: **First**, pose-guided mechanism could make the network pay more attention on identification-related areas from person images. Generally, when we need to tell whether two images are from the same person, the most natural and correct way is to focus on the person-related parts but not some confusing background areas. For example, the network should be trained to be sensitive to body parts rather than the bike or trees in background. With the help of pose cues, the network could effectively get rid of useless background noises and therefore perform better especially for industry application scenarios where we could hardly build a specific new dataset for model training and deployment. **Second**, there have been several methods that make use of pose info for part-level feature learning [5,14,15,16,17]. Most of them usually obtain the local key-point features by simply multiplying the pose heatmap with the resulting feature maps from backbone network. However, we found that such usage might be too direct and simple to make full use of pose cues. On one hand, as shown in Figure 1, global feature from backbone (e.g. ResNet50) are highly abstract that each pixel could correspond to a large area on original image and the feature of key-point pixel would easily be contaminated by unnecessary background clutters. On the other hand, due to the sparsity of heatmap, the feature that is only extracted of the highest probability point may not be robust enough especially when occlusion or posture view point variation occurs. Therefore, it is important to found a reasonable method to fully exploit the pose information. Relevant analysis and description will be shown in Section 3. **Third**, almost all of the existing part-based methods obtain the final person representation by directly concatenating the extracted features without considering the different contribution of each local feature. However, we believe that each local feature should not share the same importance with the others and that features containing more useful information deserve higher weights. For this purpose, different from some works [37,38] that use GCN for local feature fusion, we propose to use GCN to generate an independent attention weight value for each local feature, aiming to enhance the contribution of the ones with more discriminative information.

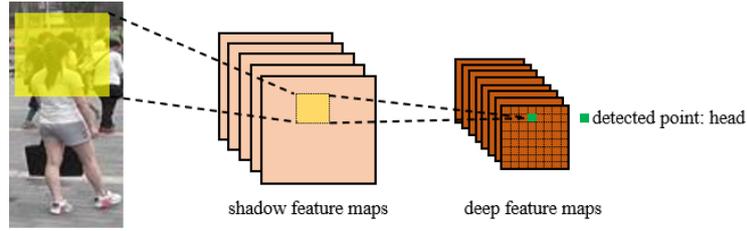

Figure 1. Illustration of reception field on deep feature maps. The detected head point on deep-level feature maps could be approximately corresponding to a relatively large area on input image, and some background noises are easily to be involved (background persons in yellow area).

Base on the analysis above, we present in this work a Pose Guided Graph Attention (PGGA) network, a multi-branch architecture consisting of one branch for global feature learning, one branch for mid-granular body features learning, one branch for fine-granular key point features learning and a GCN module. We use a pre-trained pose estimator to provide the key-point heatmaps for local feature extracting and carefully design a graph convolution layer to learn the attention weight for each extracted local feature. For global branch, we use both average and max pooling (GAP and GMP) to obtain the global vector. For mid-granular branch, we firstly generate the body attention mask matrix based on key-point locations and then add it to low-level shadow feature maps in ResNet50 backbone. Shadow feature have a smaller reception field and thus could be more suitable for applying attention mask to restrain the background noises. For fine-granular branch, we generate the fine-level attention matrices to mask the regions of interest on shadow feature maps in order to make the network focus on semantic key-point areas. For GCN module, we use a learnable adjacency matrix to learn the attention weight values based on the similarities between different local features. For inference procedure, we aggregate all the feature vectors as the final representation for person retrieval, including three global features and five local features. As will be seen in experiments, our approach shows its competitive advantages for person identity representation compared with other part-based methods. We also conduct a plenty of ablation studies and design different kinds of comparison experiments to prove its effectiveness and robustness, including the tests on holistic datasets, occluded datasets and cross-domain issue.

The main contributions of this paper can be summarized as follows:

- We design a novel pose-guided attention mechanism based on muti-branch architecture to better exploit the discriminative person representations.
- We introduce an adaptive graph attention convolution module to evaluate the contribution weight of different local features.
- Our work outperforms the state-of-the-art on several benchmarks favorably, including holistic datasets test, occlusion datasets test and cross-domain tests.

## 2. RELATED WORK

**Person reID:** With the fast development of deep learning, CNN-based feature learning [33,34,35,20,21,23] is proved to be far more effective than conventional hand-crafted methods [31,32] and thus has been widely used in person reID in recent years. CNN-based methods typically train the reID network by multi-class classification loss and ranking loss [30]. For classification loss [25], each person identity is treated as one single class and the prediction logits from fully connect layer are supervised in the training stage. For metric loss, such as contrast loss [26], triplet loss [27,28] and hard-triplet [29] loss, network is trained to maximize the distance between different identities and minimize the distance between same identities at the same time. Like most of current reID methods, this paper uses both classification and triplet loss to supervise the present network. Besides, GAN has also been proved to be effective for different kinds of reID tasks [18,19,24], especially for occluded reID and cross-domain reID. Zheng et al [19] start the first attempt to use GCN to improve the model performance by generating extra training images. Wei et al [24] proposed a Person Transfer Generative Adversarial Network (PTGAN) weaken the domain gap between source dataset and target dataset.

**Local feature exploration:** Local feature learning could enhance the discriminative person representation to a great extent by splitting the images into different parts. One branch of part-based methods is to uniformly divide the images into horizontal grids and extract the corresponding local features respectively [7,13,22,23]. In [10], Sun et al proposed a part-based convolutional network by dividing the features into several horizontal stripes and using part-level features for inference stage, which has achieved good performance in reID. In [13], Wang et al proposed a MGN model, which made use of the output of ResNet50 layer3 and trained the network with three branches. Experiment results have shown that MGN could effectively boost the performance of reID model. Generally, this branch of methods usually assumes that the body parts of particular grids on matching images are well-aligned. However, this could be problematic for real-world deployment, where the pre-trained detector tools (e.g. YOLO, SSD) might not always localize the person tightly. Another branch is to use auxiliary pose cues to capture the local features of certain body parts [5,15,16]. These methods used extra skeleton estimation model to provide human pose information. However, the combining manner they used to generate local features is to simply multiply the pose heatmap with feature maps, in which some unexpected background noises might also be involved due to the high concentration of deep-level features. Besides, there are also some other researches [11,21,2] focusing on designing attention-based networks which have been proved effective to enhance the key features of different channels or pixels. Motivated by these works mentioned above, we design a novel fusion manner to leverage the pose information by pre-trained estimator and the propose network will be detailed in section 3.

# 3. METHOD

## 3.1 Overview

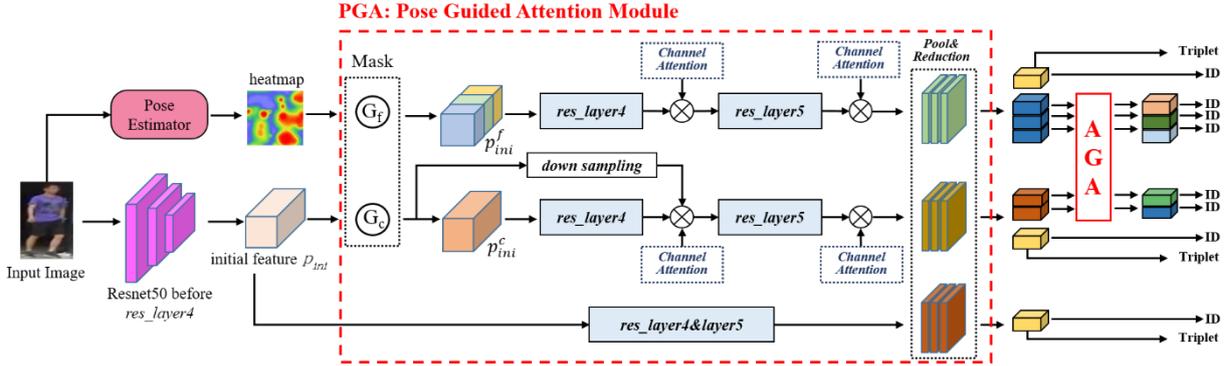

Figure 2. The pipeline of the proposed PGGA network. It mainly consists of two parts: a multi-branch backbone architecture combined with pose-guided attention mask to enable local feature learning of different granular level and a well-designed adaptive graph attention module to evaluate the contribution weights for different local features. The final representation contains three global features and five weighted local features respectively from different level. We use triplet loss combined with classification loss for training. All the features are concatenated for inference and testing.

Figure 2 shows the overview of our proposed network architecture. We design our backbone on the basis of *ResNet50* but particularly split the network into two connected parts: a shadow part consisting of the layers before *res_layer4* and a deep one consisting of three branches with each branch sharing the similar architecture with layers after *res_ layer4*. Each input person image is firstly processed by shadow part to obtain the low-level features and the output is regarded as initial feature maps for deep part network. At the same time, we extract the pose information of each input image by using a pose estimator [36], which is pre-trained on MPII [41] dataset with parameters setting fixed. In order to generate the local salient features, we introduce a Pose-Guided Attention (PGA) module based on deep part backbone to jointly extract the local features by three different levels: global feature, coarse-granular level and fine-granular level features. Subsequently, the resulting features are fed through a carefully designed Adaptive Graph Attention (AGA) module to generate the final local features of size $1\times1\times C$. Inspired by the excellent ability of Graph Convolutional Network (GCN) on node relation modeling, we aim to use AGA module to further strengthen the contribution of the ones containing more representative information. We concatenate the global vectors from PGA and all the resulting features from AGA along the channel dimension as the final representation for testing procedure. We train the whole network end-to-end by classification loss and triplet loss.

## 3.2 Pose-Guided Attention Module

Local feature learning by using part-based methods have been proved to be effective for diverse person representations. Meanwhile, pose cues could provide a relative accurate body structure information and thus has been employed by several works [5,14,16] for semantic-level local feature extracting. As mentioned in Section 1, we found from experiments that simply multiplying the pose heatmap with the output features from backbone network may not be the best way to make full use of pose cues. The reason could be summarized two-fold: 1) Features from deep layers have large reception field and some background clutters around the key-point on original image could be involved; 2) Due to the sparsity of key point probability heatmap, occlusion or different posture view point could result in unexpected representation bias of extracted semantic body parts. To alleviate these problems, we design our PGA module by combining the pose information with a multi-branch network architecture, which could effectively improve the network performance on local feature learning.

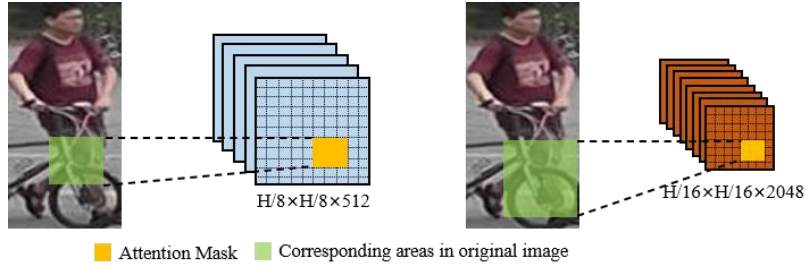

Figure 3. Illustration of the layers to choose for attention mechanism. Compared with deep features, adding attention mask on shadow feature maps of size H/8×W/8×512 could bring in less background clutters (e.g. the bike in the left green area are less involved in attention mask than the right one).

We divide the network into three different branches: one global branch to learn the global feature of person image, one coarse-granular branch to focus on the coarse areas that contains the person body and one fine-granular branch to focus on the fine-grained areas that contain the thirteen defined body parts. We split the *Resnet50* model from layer *res_conv4* and use the output feature maps of size H/8×W/8×512 as the input of PGA module, where *H* and *W* are the original image height and width. As shown in Figure 3, the reason we choose H/8×W/8 feature maps rather than H/16×W/16 final ones from *res_layer5* for applying attention mechanism is due to the considering that shadow feature maps have a relatively small reception field with more detailed information and thus the features of key-point areas could be separated more easily for better local feature learning. For the given initial feature maps, we then construct our PGA module by leveraging pose information. The pose heatmaps indicate the locations of these detected key-points, which could effectively guide the network pay more attention on human key-point areas that contain more discriminative information than other background areas.

Concretely, for a given input image of size H×W×3, we use a pre-trained pose estimator and the shadow layers before *res_conv4* to simultaneously extract the key-point heatmap and initial feature maps, which are respectively denoted by $p_{hmap}$ and $p_{ini}$. According to [36], we define thirteen semantic parts (Figure 4) to represent the human structure, including head, left shoulder, right shoulder, left elbow, right elbow, left hand, right hand, left hip joint, right hip joint, left knee, right knee, left foot and right foot. The dimensions of the mentioned variables in this paper are summarized in Table 1 in section 4.2.

For global branch, we remain the layers of *Resnet50* from *res_conv4* unchanged for global feature learning of person image. We use the combination of Global Max Pooling (GMP) and Global Average Pooling (GAP) to squeeze the feature maps to a single vector. Then the final representation $p_{global}$ is obtained by the following formulation:

$$p_{global} = R_{global}(GAP(f_{global}(p_{ini})) + GMP(f_{global}(p_{ini}))) \qquad (1)$$

where $f_{global}(*)$ is the branch network; $R_{global}$ is a subsequent sub-module to pooling layer, consisting of a *1 ×1* convolution, batch normalization (BN) and ReLU activation function layer.

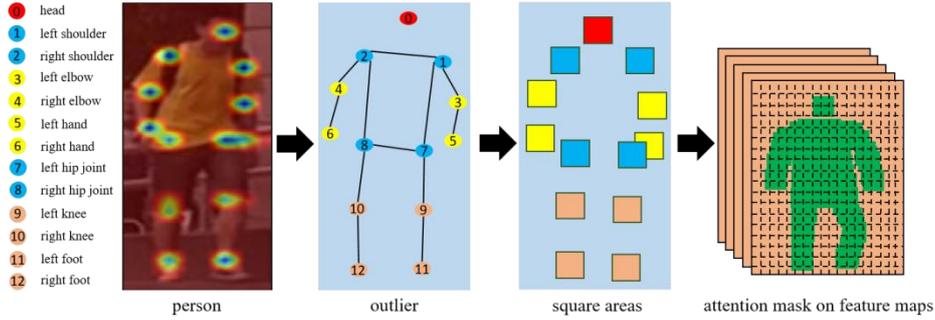

Figure 4. Illustration of the attention mask for coarse branch. The thirteen square areas are firstly generated according to the locations of detected key-points. Then we connect these squares to construct a closed attention area and obtain the mask matrix by filling with the designed coefficient. Such attention mechanism could improve perceptivity of the network on the coarse area of human body as a whole but not just on some limited parts that are vulnerable to posture variance or occlusion, such as shoes, hands and elbows.

For coarse-granular branch, we aim to make the network focus on the coarse areas that contain the human body as a whole. Generally, for a given image, the max value point on each generated heatmap indicates the location of corresponding key-point. By using all the thirteen nodes, one could build an outlier of human body on the image. $p_{hmap}$ is an array containing thirteen key-point probability matrices with value in range of (0,1) and we notice from experiments that for each matrix, only the predicted point has a high value (e.g. 0.8~1.0) while the other pixels have relatively small values (e.g. 0~0.2). That is, the heatmap is a sparse matrix where most of the elements are close to zero except for the detected key-point ones. Thus, directly multiplying the heatmap with $p_{ini}$ would inevitably results in sparse characteristic of features and severe degradation of detailed information, especially the areas around predicted point. To alleviate this problem, we tend to use the location of each max point rather than the probability value itself to construct a coarse-granular attention mask $Mask_c$ that covers the whole person body for initial feature $p_{ini}$, formulated as follow:

$$A_i = C_{area}\{(l_i^{row} - \omega, l_i^{col} - \omega), (l_i^{row} + \omega, l_i^{col} + \omega)\} \tag{2}$$

$$Mask_{row,col}^c = \begin{cases} \alpha(1-\beta), & (row, col) \in A_{outlier} \\ \beta, & else \end{cases} \tag{3}$$

where $C_{area}\{(l_1^{row}, l_1^{col}), (l_2^{row}, l_2^{col})\}$ refers to the square area determined by coordinates of upper-left point $l_1$ and lower-right point $l_2$; $(row, col)$ is the element subscript of matrix. We firstly extract a square area A centered on each single peak point by an expansion coefficient $\omega$. Then according to body structure, we connect and aggregate the area $A_i$ of different body parts to generate six closed outliers, including head, upper torso and four limbs. Then we could obtain $Mask_c$ by filling these outliers with the designed weight parameters $\alpha$ and $\beta$, which is used to indicate the contrast between attention areas and non-attention ones. Detailed description is shown in Figure 4.

Noticing that for PGA module, each branch consists of a *res_layer4* and a *res_layer5* module, and the product of $Mask_c$ and $p_{ini}$ is used as the input of this coarse branch. In order to fully leverage the pose cues, we also deploy an attention mask on the mid-result output from *res_layer4*. Since the features from *res_layer4* are more concentrated and abstract, we apply bilinear down sampling on $Mask_c$ to slightly lower the contrast degree and make its size accord with the feature size. In addition, considering that pose mask could only provide an attention area on spatial dimension, we introduce a channel-wise attention sub-module to further enhance the salient features of different channels. Formally, the features of this coarse branch of size H/16×W/16×2048 is constructed as follows:

$$p_c = Att_2(f_2(f_{ds}(Mask_c) * Att_1(f_1(p_{ini} * Mask_c)))) \tag{4}$$

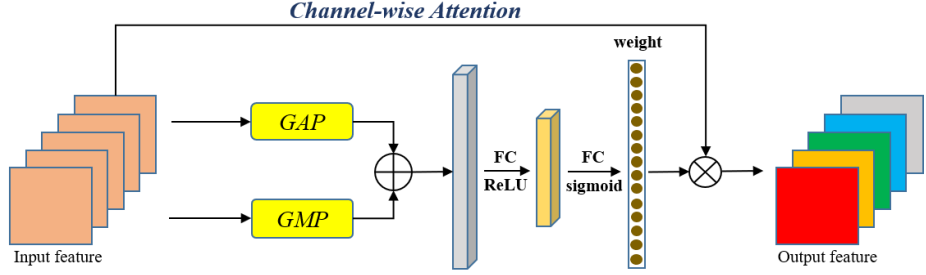

Figure 5. Illustration of channel-wise attention. Since we use attention mask to increase the feature value, using GAP combining with GMP could remain more discriminative features. The output feature is obtained by multiplying the weight value with input feature maps along channel dimension.

where $f_{ds}$ is the down sampling operation; * is element-wise product; $f_1$ and $f_2$ are *res_layer3* and *res_layer4* network, respectively. $Att$ is the channel-wise attention sub-module which is deployed twice respectively on the resulting feature of *res_layer3* and *res_layer4*. We show the structure in Figure 5, different from [40], we jointly use GAP and GMP in $Att$ module to better leverage the pose cues. In order for part-level feature learning, we split $p_c$ into two horizontal grids and extract one global with two local vector representations of size 1×1×256 by following formulations:

$$p_g^c = R_g^c(GAP(p_c) + GMP(p_c)) \tag{5}$$

$$p_{l_i}^c = R_{l_i}^c\left(GAP(p_{grid_i}^c)\right), i = 1,2 \tag{6}$$

where $R_g^c$, $R_{l_i}^c$ are reduction module sharing the same architecture with $R_{global}$; $p_{grid_i}^c$ is the partitioned feature maps from $p_c$.

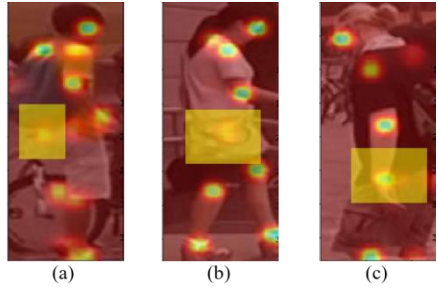

Figure 6. Illustration of occlusion and body posture variance. In picture (a), the left hand is occluded due to the side view of person image. In picture (b), the hip joints and left hand are occluded by a backpack. In picture (c), the hip joints, right arm and right hand are all occluded by left side body. Even though these key points (yellow rectangle areas) are invisible in image, the locations of them could still be predicted by pose estimator.

For fine-granular branch, we aim to make the network focus on the fine local areas of thirteen defined semantic body parts. An intuitive way is directly extracting the vector at the detected points or multiplying the heatmap with feature $p_{ini}$ and obtain the vectors of corresponding body parts by max-pooling, which is also the common usage in most of pose-based reID methods. However, preliminary experiments show that such usage might sometimes misguide the network especially when it comes to occlusion or body posture variance. For example, as shown in Figure 6 (a), the left hand is occluded by body torso, but the pose estimator could still predict its location. In this circumstance, the features of predicted hand area on $p_{ini}$ actually represent the torso but not the hand. Considering that the occlusion of certain body parts is inevitable, we believe that it is unreasonable to separately extract feature vector for each single semantic part. On the contrary, we tend to guide the network learn by itself which part to focus and obtain the local features by horizontal splitting in this branch.

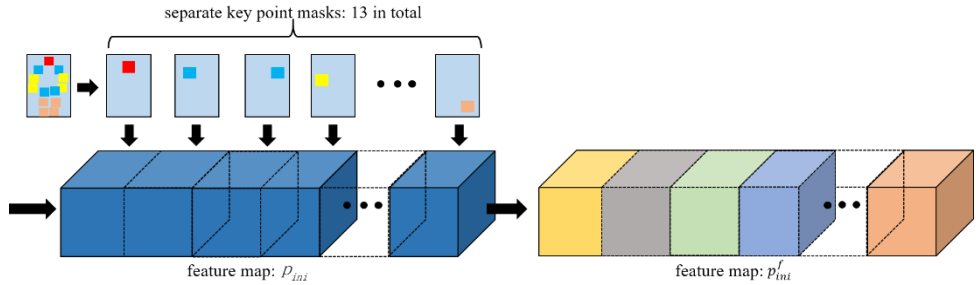

Figure 7. Illustration of the mask attention for fine granular branch. We extract thirteen different attention masks according to the locations of detected key-points, and then multiply them with the corresponding thirteen groups in channel dimension to generate the input feature of this branch.

Different from coarse-granular branch, we do not aggregate $A_i$ but directly generate a separate attention mask matrix for each predicted key-point in a similar way as Equation (3):

$$Mask_{row,col}^{f_n} = \begin{cases} conf_n * \alpha(1-\beta), & (row, col) \in A_i \\ \beta, & else \end{cases}, n = 1,2,3 \dots 13 \qquad (7)$$

where $conf_n$ is the confidence score of key-point, which could indicate the feature quality to some extent. Then we divide the initial feature $p_{ini}$ into thirteen groups in channel dimension with each group multiplying the corresponding defined mask matrix. Figure 7 shows the detail and the formulation are as follows:

$$p_{ini}^f = p_{ini} * T(Mask_{group_1}, Mask_{group_2}, \ldots, Mask_{group_n}), n = 13 \qquad (8)$$

$$Mask_{group_i} = T(Repeat(Mask_{key_i}, n_i)) \qquad (9)$$

where $p_{ini}^f$ is used as the input of *res_layer3* in this branch. We denote by T the concatenate operation in channel dimension. $Repeat(*, n_i)$ is the duplication of * by $n_i$ times and $n_i$ refers to the assigned channel numbers for each key-point mask, which is 512 exactly in total. In this way, feature maps of different channel groups are enhanced by different areas of key-points. Then we use channel-wise attention to automatically learn the weights of different channels. Similarly, we split the resulting feature maps $p_f$ into three horizontal grids, and extract one global with three local vector representations of size $1 \times 1 \times 256$ by following formulations:

$$p_f = Att_2(f_2(Att_1(f_1(p_{ini}^f)))) \qquad (10)$$

$$p_g^f = R_g^f(\text{GAP}(p_f) + \text{GMP}(p_f)) \qquad (11)$$

$$p_{l_i}^f = R_{l_i}^f\left(GAP\left(p_{grid_i}^f\right)\right), i = 1,2,3 \qquad (12)$$

In summary, by introducing pose attention, our proposed PGA module is able to fully develop the person-related features and effectively suppress those non-salient ones from background noise, which consequently leads to a more discriminative and robust person representation.

### 3.3 Adaptive Graph Attention Module

In PGA module, by taking advantage of pose information, we obtain eight feature vectors: three global ones, two coarse-granular local ones and three fine-granular local ones. A common practice for inference is to directly concatenate all these vectors in channel dimension and matching the candidate image pairs by Euclidean or Cosine distance, which is also used in this work for fair comparison with other works. In other words, each of these extracted vectors is considered to have the same contribution on similarity distance. However, since different vector contains different information accounting for person image, it is reasonable to re-evaluate the significance of each vector.

Besides, we insist that re-weighted features are robust against misalignment problem to some extent. For some datasets such as Market-1501 and DukeMTMC-reID, there are only a few samples are horizontally misaligned with others since the bounding box of the person image is annotated manually. But for real-world applications or dataset like MSMT17, bounding boxes are usually generated by object detector automatically, which may not match the person tightly, and therefore brings in unnecessary extra background noises. In such circumstance, misalignment would harm the local feature representation of matching images and thus lower the retrieval accuracy, especially for part-based methods. By re-weighting the part-level local features, the weight of misalignment part with less person relevant information would be lowered and the contribution of those parts with discriminative features would be enhanced. Thus, even though the extracted local features are not tightly aligned, the matching results would still be accurate. Detailed discussion will be shown in section 4.4.

In recent years, Graph Convolutional Network (GCN) have been applied by several works [14,37,38] for reID tasks. These works use GCN to model the spatial of temporal relations between different node features by building the adjacency matrix. However, for multi-branch networks like our PGA module, the features from different branches are not in the same feature space because of the unshared convolution kernel weight of each branch. That is, directly using GCN to generate the fused features may cause sub-optimal results, which is also proved by our preliminary experiments. Therefore, different from existing methods, we propose to apply GCN to generate the significance weight coefficient for each node features by modeling the similarity relations. Figure 8 shows the structure of our AGA module.

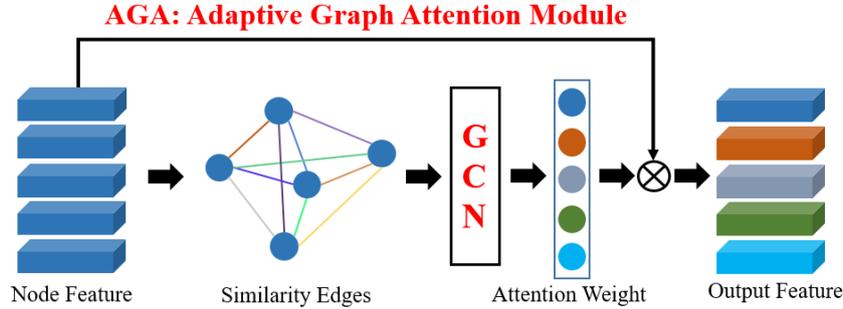

Figure 8. Structure of proposed AGA module. Different from other GCN-based methods, we use two learnable linear matrices to obtain the similarity edges. Due to the strong ability of GCN on relation modeling, the network could automatically learn the contribution weights of different input node features.

Specifically, we set the weight of all the three global vectors to 1 and use the other five local ones as input node features, denoted by $v_i \in R^{1 \times 1 \times 256}, i = 1,2,3,4,5$. Motivated by [43], we use similarity distance to generate the edges by:

$$e(v_i, v_j) = \varphi_{v_i}^T \cdot \varphi_{v_j} \tag{13}$$

where $\varphi$ represent the linear transformation of node pairs with its parameter being learnable via back propagation. Then we construct the adjacency matrix by employing L2-normalization on each row of the affinity matrix and establish the GCN module by:

$$A_{i,j} = \frac{e(v_i, v_j)}{\|e\|_2} \tag{14}$$

$$\theta = AVW \tag{15}$$

where $W \in R^{d \times 1}$, $V \in R^{d \times 5}$V and $d$ is the dimension of each node feature; θ is the resulting weight coefficient representing the contribution of each local feature. Then by multiplying it with input vectors, we finally obtain the weighted local features as:

$$Feature_{local_i} = \theta_i * v_i \tag{16}$$

### 3.4 Loss Function

We use Softmax Cross-Entropy loss (ID loss) for classification and hard-triplet loss for metric learning as the supervising function in training stage. Different from triplet loss, hard-triplet loss aims at maximizing the distance between hardest negative pair while minimizing the distance between hardest positive pair by a margin at the same time. The ID loss and hard-triplet loss in a mini-batch can be described as:

$$L_{id} = - \sum_{i=1}^{P \times K} \log \frac{e^{w_{y_i}^T x_i}}{\sum_{c=1}^{N_{id}} e^{w_c^T x_i}} \quad (17)$$

$$L_{tri} = \sum_{i=1}^{P} \sum_{j=1}^{K} \left[ \max_{p=1,2...K} \|A_i - P_i\|_2 - \min_{\substack{n=1,2...K \\ j=1,2...K \\ j \neq i}} \|A_i - N_i\|_2 + \gamma \right]_+ \quad (18)$$

In equation (x), $N_{id}$ denotes the number of identities in the whole training set; $W$ denotes the weight parameters of fully connected (FC) layer; $y_i$ is the ground truth identity of input image and $x_i$ is the output of FC layer for ith identity. In equation (x), $A_i$, $P_i$ and $N_i$ are the feature vectors of anchor, positive and negative samples respectively; $\gamma$ is the margin to the distance between negative and positive pair. For better performance, we set the bias terms in linear multi-class classifiers to zero. We employ the ID loss on global features and all the part-level local feature vectors. Considering that part-level local features are not suitable for selecting out the hardest positive and negative pairs, triplet loss is only deployed on global features. The total loss for training stage is constructed by a weight parameter $\tau$ as follow:

$$L = L_{tri} + \tau * L_{id} \quad (19)$$

## 4. EXPERIMENT

**4.1 Dataset**

We conducted experiments on three holistic datasets and two occluded datasets: 1) The **Market-1501** [48] dataset contains 1,501 person identities and 32,217 images from 6 different cameras. The whole dataset is split into training set and testing set. Training set consists of 12,936 images of 751 identities. Testing set consists of 3,368 query images and 19,732 gallery images of 750 identities. 2) The **DukeMTMC-reID** [49] dataset contains 1,812 person identities and 36,411 images from 8 high resolution cameras without any cross-view. Training set consists of 16,522 images of 702 person identities which are randomly selected from dataset. Testing set consists of 2,228 query images and 17,661 gallery images of the other 702 persons in total. 3) The **MSMT17** [51] dataset contains 4,101 person identities and 126,441 images from 15 different cameras. Training set consists of 32,621 persons of 1,041 identities. Testing set consists of 93,820 persons of 3,060 identities. The bounding boxes are labeled by faster RCNN automatically. For testing, 11,659 boxes are randomly selected as query while the other 82,161 ones regarded as gallery. 4) The **Occluded-Duke** [18] dataset is selected from DukeMTMC-reID, containing 15,618 training images, 17,661 gallery images and 2,210 occluded query images. 5) The **Occluded-reID** [55] dataset contains 2,000 images of 200 occluded persons, captured by mobile camera. Each person has five full-body images and five occluded images.

## 4.2 Implementation Details

Table 1. Dimensions of network variables.

| Description | Variable | Dimension (Height×Width×Channel) |
|---|---|---|
| Input image | $P_{input}$ | H×W×3 |
| Detected heatmap | $P_{hmap}$ | H/8×W/8×13 |
| Feature maps resulting from *res_layer2* | $P_{ini}$ | H/8×W/8×512 |
| Input feature maps for coarse branch | $p_{ini}^c$ | H/8×W/8×512 |
| Input feature maps for fine branch | $p_{ini}^f$ | H/8×W/8×512 |
| Resulting feature vector from global branch | $P_{global}$ | 1×1×256 |
| Resulting feature vectors from coarse branch | $p_g^c, p_{l_i}^c$ | 1×1×256 |
| Resulting feature vectors from fine branch | $p_g^f, p_{l_i}^f$ | 1×1×256 |

We summarize the dimensions of different variables in Table 1. The implementation settings are listed below. Resnet50 backbone network is initialized by the weights pre-trained on ImageNet. The FC and pooling layers of standard Resnet50 network are removed and the last stride of *res_layer5* is set to 1. We use HRNet [36] as pose estimator with fixed pre-trained weights on MPII [41] dataset. Each input image is resized to 384×128. We deploy random horizontal flipping and random erasing as data augmentation methods in training stage. The initial learning rate is set to 0.01 and 0.001 for the backbone and other parts. The margin parameter for hard triplet loss function is set to 1.2. SGD optimizer is used with momentum set to 0.9 and weight decay factor set to 0.0005. The expansion coefficient ω is set to 2. The mask weight parameter α and β are set to 2 and 0.5 respectively. The loss weight parameter $\tau$ is set to 2. In order to fully demonstrate the effectiveness of our network, we do not apply some training tricks that could lead to slight advance on testing results, such as warmup strategy, label smooth, circle loss etc. In inference stage, all the vectors are concatenated to represent the person feature for retrieval. Our work is implemented on four NVIDIA RTX 1080ti GPUs based on PyTorch framework (version 1.6.0). All the experiments on different datasets follow the same settings mentioned above.

## 4.3 Comparison with State-of-the-Art Methods

**Results on holistic datasets.**

We show in Table 2 the quantitative comparisons (without reranking) of our network with the state-of-the-art works. We mainly choose part-based, attention-based and GCN-based methods that are proposed in recent three years to make the comparisons. We use mean average precision (mAP)(%) and rank-1 accuracy (%) to evaluate the network. As shown, our work gives the best results on three datasets in terms of mAP and rank-1. We achieve mAP of 90.5% and rank-1 accuracy of 96.5% on Market-1501, mAP of 80.5% and rank-1 accuracy of 91.1% on DukeMTMC-reID, mAP of 61.9% and rank-1 accuracy of 84.8% on MSMT17. Particularly, compared with other existing pose-based works, such as [14, 16], our method is the only one that achieves the SOTA results on holistic datasets.

Table 2. Performance caparisons with SOTA on holistic datasets.

| Method | Market-1501 | | DukeMTMC | | MSMT17 | |
|---|---|---|---|---|---|---|
| | mAP | Rank-1 | mAP | Rank-1 | mAP | Rank-1 |
| MGN[13] | 86.9 | 95.7 | 78.4 | 88.7 | - | - |
| PCB+RPP[7] | 81.6 | 93.8 | 69.2 | 83.3 | 40.4 | 68.2 |
| RelationNet[39] | 88.9 | 95.2 | 78.6 | 89.7 | - | - |
| RGA-SC[46] | 88.4 | 96.1 | - | - | 57.5 | 80.3 |
| SCAL[47] | 89.3 | 95.8 | 79.1 | 88.9 | - | - |
| SONA[48] | 88.83 | 95.58 | 78.18 | 89.55 | - | - |
| HOreID[14] | 84.9 | 94.2 | 75.6 | 86.9 | - | - |
| MGCAM[49] | 74.33 | 83.79 | - | - | - | - |
| Pyramid[50] | 88.2 | 95.7 | 79.0 | 89.0 | - | - |
| DenseS[51] | 87.6 | 95.7 | 74.3 | 86.2 | - | - |
| ABD[52] | 88.28 | 95.6 | 78.6 | 89.0 | 60.8 | 82.3 |
| SCSN(3stages)[53] | 88.5 | 95.7 | 79.0 | 90.1 | 58.0 | 83.0 |
| **ours-PGGA** | **90.5** | **96.5** | **80.5** | **91.1** | **61.9** | **84.8** |

**Results on occluded datasets.**

We show in Table 3 the experiment results on occluded datasets. For Occluded-Duke, the model is pre-trained on DukeMTMC-reID. For Occluded-reID, the model is pre-trained on Market-1501. We take all the occluded person images as query set and the full body images as gallery set on both two datasets. It should be noticed that some of the works listed are specially designed for occluded reID task [16,54,5,14] while our network is designed for general purpose. Comparison results show that our method significantly outperforms the existing SOTA works on Occluded-Duke with a performance of 58.8%(mAP)/70.9%(Rank-1) and achieve a very close performance to SOTA results on Occluded-reID. Note that images in Occluded-Duke are much more difficult to retrieve due to the large pose variations, occlusions and confusing noises, which further demonstrates the advantage of our work.

Table 3. Performance caparisons with SOTA on occluded datasets.

| Method | Occluded-Duke | | Occluded-reID | |
|---|---|---|---|---|
| | mAP | Rank-1 | mAP | Rank-1 |
| PCB [7] | 33.7 | 42.6 | 38.9 | 41.3 |
| PGFA [16] | 37.3 | 55.1 | - | - |
| Teacher-S [54] | 22.4 | 18.8 | 59.8 | 55.0 |
| PVPM [5] | 29.2 | 51.5 | 61.2 | 70.4 |
| HOreID [14] | 43.8 | 55.1 | **70.2** | **80.3** |
| DSR [45] | 30.4 | 40.8 | 62.8 | 72.8 |
| **Ours-PGGA** | **58.8** | **70.9** | 70.1 | 79.5 |

### 4.4 Discussion

**Ablation Study.**

We conduct several ablation experiments on the different components of our model in Table 4. We respectively denote the global branch, coarse-granular branch, fine-granular branch, pose mechanism, channel wise attention sub-module and proposed GCN module by G, C, F, POSE, ATT and AGA. Our baseline is a Resnet50 model followed by a reduction layer to squeeze the feature dimension to 256. The 2$^{nd}$ and 3$^{rd}$ rows are designed to separately evaluate the performance of each single branch. The 4$^{th}$ to 6$^{th}$ row are designed to evaluate the effectiveness of the pose attention and channel attention. Specifically, for the 4$^{th}$ row, the absence of POSE means that the network of C-branch and F-branch are constructed without pose mask but only consists of the original *res_layer4* and *res_layer5*. The final row is designed to evaluate the proposed AGA module. All the experiments follow the same settings mentioned in section 4.2. From the results, we can observe that: 1) Either C-branch or F-branch significantly outperforms the baseline. For example, F-branch brings in an improvement of 5.5%(mAP)/2.2%(Rank-1) on Market-1501 and 7.1%(mAP)/3.5%(Rank-1) on DukeMTMC-reID dataset. 2) The results in 5$^{th}$ row demonstrate the effect of our proposed pose attention mechanism. With the help of key-point locations, the network could pay more attention on person-related areas. On DukeMTMC-reID, this gives a performance gain of 1.7%(mAP)/0.9%(Rank-1). On Market-1501, even thought the performance is already very high, it still brings in an improvement of 1.9%(mAP)/1.2%(Rank-1). 3) Channel attention is proved to be an effective complement module for our pose mechanism. For example, given a feature of size $48\times16\times512$, we apply the pose masks on all the 512 maps. But we notice that some of these maps contain salient feature patterns in mask areas while some others not. Thus, our designed ATT module could help to enhance the maps with more useful information and suppress the ones with background clutters. 4) The results in final row indicate the effectiveness of our proposed AGA module, which respectively brings in a gain of 0.5%(mAP)/0.4%(Rank-1) on Market-1501 and 0.4%(mAP)/0.6%(Rank-1) on DukeMTMC-reID. Detailed discussion with visual proof is shown in latter paragraph.

Table 4. Quantitative comparison of different network components.

| Component | POSE | ATT | AGA | F-dim | Market-1501 | | DukeMTMC | |
|---|---|---|---|---|---|---|---|---|
| | | | | | mAP | Rank-1 | mAP | Rank-1 |
| Baseline: single-G | × | × | × | 256 | 83.4 | 92.7 | 72.1 | 85.8 |
| G + C | √ | √ | × | 1024 | 88.5 | 94.8 | 78.6 | 89.4 |
| G + F | √ | √ | × | 1280 | 88.9 | 94.9 | 79.2 | 89.3 |
| G + C + F | × | × | × | 2048 | 87.7 | 94.5 | 78.1 | 88.8 |
| G + C + F | √ | × | × | 2048 | 89.6 | 95.7 | 79.8 | 89.7 |
| G + C + F | √ | √ | × | 2048 | 90.0 | 96.1 | 80.1 | 90.5 |
| PGGA: G + C + F | √ | √ | √ | 2048 | 90.5 | 96.5 | 80.5 | 91.1 |

**Pose Attention Mechanism.**

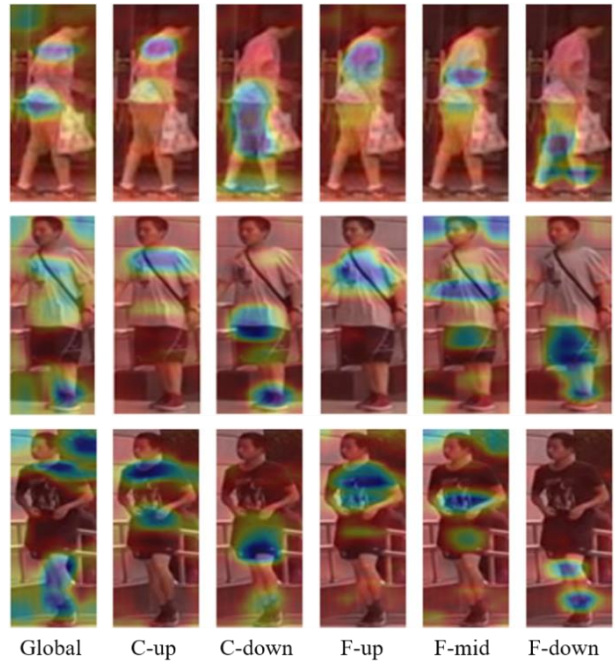

Figure 9. Class activation map extracted from *res_layer5_conv3* of each branch by Grad-CAM [55]. Images are sampled from Market-1501 dataset. 1st column: activation heatmaps from global branch; 2nd to 3rd column: activation heatmaps of the two horizontal grids from coarse-granular branch; 4th to 6th column: activation heatmaps of the three horizontal grids from fine-granular branch. Dark color denotes higher value.

In order to evaluate the effectiveness of our proposed pose attention mechanism, we visualize the class activation maps from *res_layer5_conv3* of each branch by using Grad-CAM [55]. Figure 9 shows some typical examples selected from Market-1501 dataset. For global branch, we can notice that some salient features are easily ignored and sometimes the irrelevant background clutter could mislead the network. In the global map of the 2nd row, only the shoes are highly preferred and the information from shirt, pants or legs is commonly ignored. In the global case of the 3rd row, the area next to the head is actually background noise that contain no useful cues about identity for pedestrian but it is still be highly concentrated on. For local branches, we can see from the figure that by taking advantages of pose information, our proposed PGA module could fully exploit the local semantic features and filter out most of the confusing background noises at the same time. For example: 1) In the cases of the 1st row, C-branch mainly focus on upper and lower torso areas, such as the T-shirt and pant, while the responses of F-branch concentrates more on body parts such as the elbow, knees and shoes. 2) On global response map of the 3rd row, by applying pose attention masks, the misleading area at the right of person's head is effectively filtered in C-branch and F-branch. Therefore, the visualization proofs indicate that our proposed PGA module allows the network to fully mine the diverse and discriminative local features on different branches.

**Adaptive Graph Attention Module.**

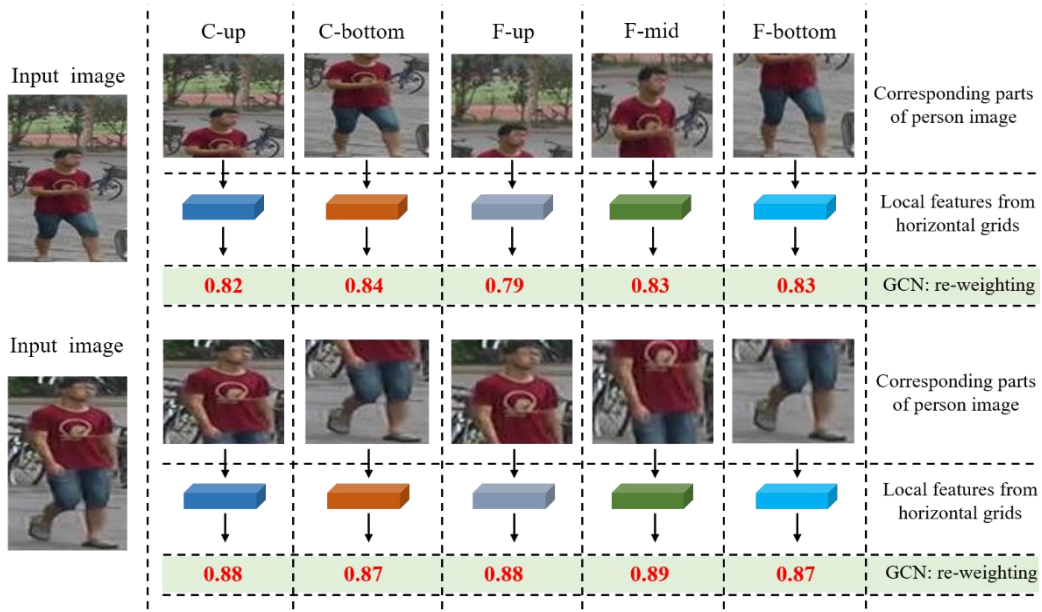

Figure 10. Samples of weight coefficient in proposed AGA layer. The two images are from a same person identity of Market1501 dataset. Weight coefficients shown in red color are extracted and printed out from testing procedure. Each smaller part picture is cut from the input image according to each split horizontal grid with reception field taken into consideration.

We show in Figure 10 some weight coefficient results extracted from testing procedure. We select two typical person images of a same identity as inputs to illustrate the rationality of our proposed AGA layer. That is, the first image contains more background interference and the person body in two images are not aligned tightly. Based on a large amount of observation experiments, we found that: 1) Features containing less discriminative information tend to be assigned a lower weight. For example, in the first image, obviously there are lots of background noises (e.g. bikes and trees) in the pictures of C-up and F-up parts due to the imprecise annotation bounding box and the resulting weights of them are consequently lower (0.82/0.79), while the other parts that contain useful features (e.g. holistic shirt and pants) are assigned higher weights (0.84/0.83/0.83). 2) Features from high-quality image samples are usually learned to have higher contribution weights. In particular, high-quality sample refers to the person image in which the bounding box is perfectly annotated, less background clutters exists and no occlusion occurs. For example, the average feature weights of the second image is larger than the first one. 3) The values of weights are usually trained to be in an approximate range of (0.75,0.95). We infer that such phenomenon is to some extent related to the number of horizontal grids. We split the feature maps into two/three stripes in our method and each grid feature still contains useful information more or less, even though the image sample are pictured with a bad misalignment problem.

**Cross Domain Test.**

Table 5. Quantitative comparison of different network components.

| Method | Market-1501→DukeMTMC-reID | | | | DukeMTMC-reID→Market-1501 | | | |
|---|---|---|---|---|---|---|---|---|
| | mAP | Rank-1 | Rank-5 | Rank-10 | mAP | Rank-1 | Rank-5 | Rank-10 |
| BagTricks [29] | 26.7 | 44.1 | 59.7 | 65.7 | 29.2 | 57.4 | 73.8 | 78.7 |
| RelationNet [39] | 25.7 | 42.8 | 59.1 | 65.0 | 34.0 | 61.2 | 76.7 | 82.4 |
| PCB [7] | 26.5 | 46.1 | 60.9 | 66.6 | 36.0 | 63.3 | 79.2 | 84.3 |
| MGN [13] | 31.8 | 50.0 | 65.5 | 70.2 | 40.0 | 68.2 | 81.6 | 86.2 |
| EANet [60] | 36.0 | 56.1 | - | - | 35.8 | 66.1 | - | - |
| SPGAN+LMP [56] | 26.2 | 46.4 | 62.3 | 68 | 26.7 | 57.7 | 75.8 | 82.4 |
| HHL [57] | 27.2 | 46.9 | 61.0 | 66.7 | 31.4 | 62.2 | 78.8 | 84.0 |
| PGGA w/o pose | 32.2 | 50.2 | 66.9 | 71.2 | 39.9 | 66.6 | 80.3 | 85.4 |
| PGGA with pose | **38.2** | **58.3** | **71.1** | **76.2** | **43.3** | **72.3** | **84.5** | **89.1** |

Cross-domain person retrieval has always been a challenging task in which the model trained on a certain dataset would suffer a significant drop on another dataset. Such phenomenon is due to the data-bias existing between source and target datasets. For example, images from Market1501 and DukeMTMC datasets are captured under different camera environment factors including light, camera resolution, proportion of background noise, etc. Thus we conduct experiments to evaluate the performance of our method on cross-domain reID task to show its advantage, the results are shown in Table 5. "Market1501→DukeMTMC" in Table 5 means that the network is trained on Market1501 and directly test on DukeMTMC dataset without extra training. According to paper [29], we remove Random Erase Augmentation (REA) since it may lead to a severe degradation on cross-domain performance. The methods are categorized into two groups: 1) We re-implemented four typical part-based methods, including BagTricks, PCB, MGN and RelationNet. 2) EANet, SPGAN+LMP and HHL are three typical unsupervised domain adaptive (UDA-based) methods proposed in recent years, which employed unsupervised learning.

From the experiments, we can observe that: 1) The effective usage of pose cues makes our PGGA model largely outperform other part-based works. 2) The comparison with the second group indicates the strong generalization ability of our method. Even though our work is not designed for cross-domain task on purpose, our PGGA model could still achieve a higher performance over most of UDA-based works. It should be noticed that the methods in second group apply unsupervised training on target dataset, while we directly transfer the model for test without any extra training procedures. In addition, for fair comparison, we do not take into consideration in this paper the works that employ clustering methods to generate pseudo labels for target domain training, such as UDAP [58] and AD-Cluster [59]. 3) We separately evaluate the effect of our pose attention mechanism in the third group. We found from experiments that the commonality and high accuracy of pose detector could effectively help to alleviate the gap-bias between source and target domain. That is, by such priori knowledge, the key parts of person in a target domain image could be enhanced and thus lead to a more discriminative feature representation.

**Performance of Local Feature.**

Table 6. Quantitative comparison of each extracted local feature on Market1501 dataset.

| Description | Feature | w/o POSE | | with POSE | |
|---|---|---|---|---|---|
| | | mAP | Rank-1 | mAP | Rank-1 |
| Local feature from up-grid of C-branch | $p_{l_1}^c$ | 61.3 | 79.3 | 68.9 | 85.0 |
| Local feature from bottom-grid of C-branch | $p_{l_2}^c$ | 66.0 | 81.4 | 70.4 | 88.5 |
| Local feature from up-grid of F-branch | $p_{l_1}^f$ | 52.2 | 74.8 | 65.0 | 86.3 |
| Local feature from mid-grid of F-branch | $p_{l_2}^f$ | 68.9 | 80.8 | 74.8 | 89.9 |
| Local feature from bottom-grid of F-branch | $p_{l_3}^f$ | 54.7 | 73.3 | 63.5 | 79.5 |

We conduct quantitative comparisons to demonstrate the effectiveness of our proposed pose mechanism on each separate local feature. Similar with the second group experiments in Table 4, we respectively train the whole network with and without using pose attention. Other training details follow the same settings mentioned in section 4.2. For a given query image, instead of concatenating all the local features, we use each single one of them as person representation to match the person image pairs and report their performance on Market1501 dataset in Table 6. From this table, we can observe that: 1) Our proposed pose attention mechanism could drastically improve the capability of discriminative local feature learning. It could bring in a maximum gain of 12.8% and 11.5% on mAP and Rank-1 accuracy, respectively. 2) The local feature from mid-grid of F-branch gives the best results, which suggests that the middle part (commonly body torso) of a person image usually contains more discriminative features rather than the other parts. 3) The 3[rd] and 5[th] rows indicates that our pose mechanism could perfectly lead the network locate and focus on local person-related features, especially for the situations where the image part contains more confusing background interference. For example, the upper and bottom parts of a person image are usually supposed to contain less useful information (e.g., simply head and legs) and more background noises. By applying pose attention, the performance of up-grid and bottom-grid features from F-branch are largely improved over 10% on both mAP and Rank-1 accuracy, which strongly demonstrates the ability of our pose attention on hard feature mining.

**Impact of Splitting Grid Number.**

Table 7. Comparison of different grid number settings on Market-1501.

| Global features | Grid Number | | Total features | F-dim | Market-1501 | |
|---|---|---|---|---|---|---|
| | C-branch | F-branch | | | mAP | Rank-1 |
| 3 | 2 | 2 | 7 | 1792 | 90.0 | 95.9 |
| 3 | 2 | 4 | 9 | 2304 | 89.7 | 95.6 |
| 3 | 2 | 5 | 10 | 2560 | 88.4 | 94.9 |
| 3 | 2 | 6 | 11 | 2816 | 88.6 | 94.8 |
| 3 | 3 | 3 | 9 | 2304 | 89.9 | 96.1 |
| 3 | 3 | 4 | 10 | 2560 | 90.1 | 96.0 |
| 3 | 3 | 5 | 11 | 2816 | 89.4 | 95.5 |
| 3 | 3 | 6 | 12 | 3072 | 88.3 | 94.5 |
| 3 | 4 | 4 | 11 | 2816 | 89.8 | 95.7 |
| 3 | 4 | 5 | 12 | 3072 | 89.3 | 95.4 |
| 3 | 4 | 6 | 13 | 3328 | 88.1 | 94.8 |
| **3** | **2** | **3** | **8** | **2048** | **90.5** | **96.5** |

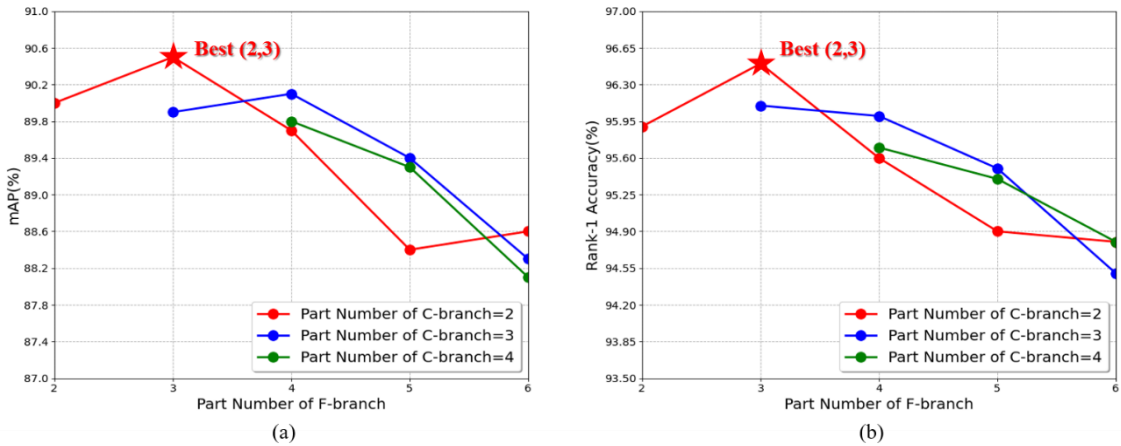

Figure 11. Graphic results of mAP (a) and Rank-1 accuracy (b) comparison with different splitting strategy on Market-1501 dataset. The best point is highlighted by red star.

We conduct several experiments to evaluate the influence caused by different part number settings. Quantitative results on different datasets are shown in Table7 and Figure 11. We denote by $N_C$ and $N_F$ the granularity of C-branch and F-branch. We set $N_F$ with a maximum of 6 and a minimum value equal to $N_C$ in each separate experiment. As can be seen from Figure 11, on one hand, the performance reaches the peak when $(N_C, N_F)$ arrives (2, 3). For a given $N_C$, with the differential $(N_F - N_C)$ increases, the performance are getting worse and we infer that the optimal constraint condition should be $N_F - N_C = 1$. On the other hand, it could be noticed from Table 7 that the dimension of the representation vector increases as the total splitting number gets higher, which to some extent leads to a negative influence on model performance. We suggest that this phenomenon might be drawn by two reasons: 1) The feature of each grid may not be able to contain enough semantic information since the feature map is "over-fine-grained" under a large splitting number setting. 2) Considering that we establish the final representation vector by concatenating all the separate features in channel dimension, using too many local features will inevitably weaken the contribution of global features. For example, in the 11$^{th}$ row of Table 7, the number of local feature and global feature is 10 and 3, respectively. Obviously, the local features will occupy a much larger proportion than global features during the calculation of Euclidean distance for matching the image pair, and consequently lower the weight of some useful semantic information contained in global features.

## 5. CONCLUSIONS

In this paper, we have presented a pose guided graph attention network, which is a multi-branch architecture consisting of three separate branches designed on purpose. We carefully design a PGA module and an AGA module to fully exploit the pose information of human body. Experiments results demonstrate the effectiveness of our approach on discriminative feature learning and we show that our model achieves state-of-the-art performances on several mainstream evaluation datasets. We also conduct a plenty of ablation studies and design different kinds of comparison experiments for our network to prove its commonality and robustness, including occluded datasets and cross-domain testing. Furthermore, our unique way of utilizing pose information could be easily extended to some existing state-of-the-art methods and can facilitate the reID research progress for both industry and academic community.